\documentclass{article}
\UseRawInputEncoding
\usepackage[preprint]{neurips_2021}
\usepackage[utf8]{inputenc}
\usepackage{amssymb}
\usepackage{amsmath}
\usepackage{mathtools}
\usepackage{amsthm}
\usepackage{wrapfig}
\usepackage{natbib}
\setcitestyle{numbers,square,citesep={,}}

\usepackage{algpseudocode}
\algnewcommand{\func}[1]{\textsc{#1}}

\usepackage[T1]{fontenc}    
\usepackage{hyperref}       
\usepackage{url}            
\usepackage{booktabs}       
\usepackage{amsfonts}       
\usepackage{nicefrac}       
\usepackage{microtype}      
\usepackage{xcolor}         

\usepackage{algorithm} 
\usepackage{algorithmicx}

\title{Real-time Drift Detection on Time-series Data}
\author{{Nandini Ramanan \,\,\,\, Rasool Tahmasbi \,\,\,\, Marjorie Sayer} \\
{\textbf{Deokwoo Jung} \,\,\,\,\,\, \textbf{Shalini Hemachandran} } \\
{\textbf{Claudionor Nunes Coelho Jr}} \\\\
{Advanced Applied AI Research} \\
{Palo Alto Networks} \\
{https://www.paloaltonetworks.com/}
}

\begin{document}

\maketitle

\begin{abstract}
Practical machine learning applications involving time series data, such as firewall log analysis to proactively detect anomalous behavior, are concerned with real time analysis of streaming data. Consequently, we need to update the ML models as the statistical characteristics of such data may shift frequently with time. One alternative explored in the literature is to retrain models with updated data whenever the model’s accuracy is observed to degrade. However, these methods rely on near real-time availability of ground truth, which is rarely fulfilled. Further, in applications with seasonal data, temporal concept drift is confounded by seasonal variation. In this work we propose an approach called \underline{U}nsupervised \underline{T}emporal \underline{D}rift \underline{D}etector or \texttt{UTDD} to flexibly account for seasonal variation, efficiently detect temporal concept drift in time series data in the absence of ground truth, and subsequently adapt our ML models to concept drift for better generalization. 
\end{abstract}

\section{Introduction}
\label{sec:intro}

In statistical machine learning, drift detection is defined as the detection of the deterioration of model performance, requiring users to retrain the model based on new inputs. As we move from a model-centric to a data-centric approach, 
users are more and more using pre-defined pipelines with large amounts of data. In this scenario, we expect model evaluation to play an important role in determining when model performance  deteriorates, as this is ultimately related to training cost in the cloud. Usually, this deterioration is detected at the input level, as a deviation from input parameters used at training time is the first indication that previous model training may no longer be valid. 



We are particularly interested in detecting model drift in temporal series (one where a portion of the inputs refer to previous values of a variable we want to forecast or to detect anomalies), where the machine learning model is used to estimate a possibly non-linear function of previous $p$ instances, such as given by Formula~\ref{temporal}, where $y_i$ and  $x_j$ are referred usually as an endogenous variable and an exogenous variable respectively.


\begin{equation}
    y_t = f(y_{t-p:t-1}, x_{t-p:t})
    \label{temporal}
\end{equation}

\begin{wrapfigure}{r}{0.40\textwidth}
\vspace{-3.0em}
    \begin{center}
    \includegraphics[width=0.40\textwidth]{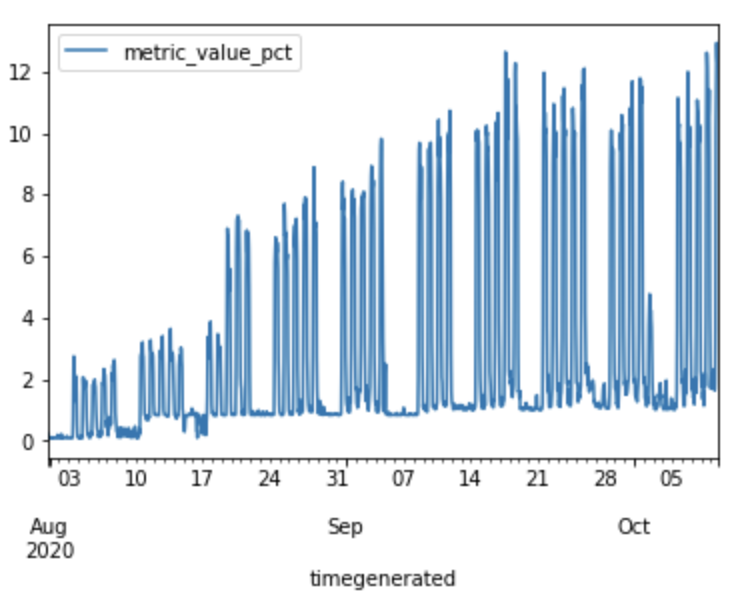}
    \caption{Synthetic time-series data with seasonality/cycles between the month of Aug to Oct 2020}
    \label{fig:train}
    \end{center}
    \vspace{-4.0em}
\end{wrapfigure}

When doing predictions or detecting anomalies in time series, one needs to consider seasonality and cycles, which may present themselves in a non-linear relation, and where their relation may not be known in a general case, such as in the example presented in Figure~\ref{fig:train}.  In this case, in order to detect drift, we need to consider the unknown effects of the seasonality and cycles, as it can be easily shown in this figure that the model contains two behaviors - one during the week, and another one during the weekends.

This paper proposes an algorithm \underline{U}nsupervised \underline{T}emporal \underline{D}rift \underline{D}etector or interchangeably referred as \texttt{UTDD} to address the problem of modeling seasonalities and cycles in a generic way so that we can detect the drift in time series variables by detecting anomalies in their behavior, after we remove the seasonalities and cycles. We consider a generic framework that accommodates unknown relations and non-linear relations as well.

This paper is organized as follows. In the next section, we present related work to drift detection, including common techniques used in drift detection. Then, we present our work to detect model deterioration based on detecting anomalies in previous variables once we remove seasonalities and cycles in a black-box way. We do this using artifacts of Boosted Embeddings. Then, we present results and conclusions.

\section{Related Work}
\label{sec:related}
Near-limitless volumes of data in temporal order, which are referred to as data streams, are generally nonstationary as the characteristics of data evolve over time. This phenomenon is called concept drift, and is an issue of great importance in the literature, because it makes models obsolete by decreasing their predictive performance. In the presence of concept drift, it is necessary to adapt to changes in data to build more robust and effective classifiers.

Gemaque \cite{driftdetection2020} presented a comprehensive overview of approaches that tackle concept drift in classification problems in an unsupervised manner. Brockhoff et.al \cite{brockhoff2020time} used Earth Mover’s Distance, a measure of the distance between two probability distributions, to detect the drift. In 2020, Liu \cite{liu2020concept} proposed a cluster-based histogram, called equal intensity k-means space partitioning (EI-kMeans). Pinagé \cite{pinage2020drift} proposed a semi-supervised drift detector that uses an ensemble of classifiers based on self-training online learning and dynamic classifier selection. Xuan \cite{xuan2020bayesian} presented a Bayesian non-parametric unsupervised concept drift detection method based on the Poly tree hypothesis test. The basic idea is to decompose the underlying data distribution into a multi-resolution representation that transforms the whole distribution hypothesis test into recursive and simple binomial tests.

On the one hand, a high accuracy detection approach usually requires labeled data, possibly involving high cost for labeling. On the other hand, a variety of methods have been devoted to the topic of concept drift detection with unlabeled data, but these approaches often are most suited for only a subset of the concept drift types. Hu et. al. \cite{hu2020no} did a survey to present these methods, categorize them and give recommendations of usage based on their behaviors under different types of concept drift. Recently, Gözüaçık and Can \cite{gozuaccik2021concept} presented an implicit (unsupervised) algorithm called One-Class Drift Detector (OCDD), which uses a one-class learner with a sliding window to detect concept drift. 
\section{Temporal Drift Detection}
\label{sec:tdd}
\noindent\paragraph{Concept Drift in Time-series Data}
Concept drift is a phenomenon in which the statistical properties of a domain change over time in an arbitrary way.
Given a set of $k$ samples from a joint distribution over a time interval $[t_1,t_k]$,
\begin{gather}
    \mathcal S = \{\mathcal D_{t_1}, \hdots, \mathcal D_{t_k}\} \sim F_{[t_1,t_k]}(X,y)
\end{gather}
where dataset $\mathcal D_{t_i}$ is comprised of both a feature vector $X_{t_i}$ and label $y_{t_i}$ such that $\mathcal D_{t_i} = (X_{t_i}, y_{t_i})$ and the notation $F_{[t_1,t_k]}$ making the connection between a joint distribution $F$ and the time interval $[t_1,t_k]$ explicit. Concept drift  can then be defined as,
\begin{gather}
    F_{[t_1,t_k]}(X,y) \neq F_{t_{[k+1, \infty]}}(X,y), \text{   \,\,or equivalently,} \\
    \exists t_j: Pr_{t_j}(X, y) \neq Pr_{t_j+1}(X,y)
\end{gather}
    
In this paper we study unsupervised model adaptation under concept drift. To be able to detect drift in the underlying concept based on the variables X alone,
we must observe some drift or changes in the feature distribution $F_{[t_1,t_k]}{[X]}$ with time.

\noindent\paragraph{Boosted Embeddings}
In this section we leverage embeddings to learn seasonality (e.g., daily, weekly, monthly) or unknown cycles by multiple categorical features. Let's assume $\theta^{T}$ to capture time-categorical features (e.g., months of the year, days of the week, and hours of the day) and $\theta^{I}$ to represent other independent categorical features. 
Let assume that a model trained is trained at at $t_1,t_2,... t_m$ where $t_{m-1} < t_{m}$.
Then, the model trained at time $t_m$ can be formulated as $f^m := f^m_{emb} (\mathbf{x} ; \theta_m^{T}, \theta_m^{I}) + f^m_{res} (\mathbf{x})$ where $f_{emb}$ is embedding model and $f_{res}$ is a residual model.
If the model drift occurs at time $t_{m+1}$,  it will incur a large $f^m_{res} (\mathbf{x})$ that will be captured by our \texttt{UTDD} algorithm. We leveraged the \textit{DeepGB} algorithm in \cite{karingula2021boosted}, where gradient boosting trains several simple models sequentially. The key idea of boosting is that each subsequent model trains only on the difference of the output and previous model, to leverage each model's strengths and minimize regression error. Their approach is conducting gradient boosting to fit weak learners on residuals to improve the previous models. They propose a loop wherein, at each iteration, they freeze the previous embedding, and add the new embedding to the sequence of models. Such that, $f^m = [e^m_1, \ldots, e^m_L, r^m].$
where $e_i, i=1,\cdots, L$ are embedding models to capture categorical data, and $r$ represents the residual model.
The summary of their boosted embedding approach is presented in Algorithm \ref{algo:BEA}.
\begin{algorithm}
\begin{algorithmic}[1]
\small
\Function{BoostedEmbeddings}{$\mathbf{X}$,$N$}
\State where $\mathbf{X}$ = $((t_n, \mathbf{y}_n))^N_{n=1}$
\State $f^m = []$
\State $F_0 := y$
\For {$1\leq l \leq L+1$:} \Comment{iteration over the embedding models}
\State $e_l.fit(t,F_{l-1})$ \Comment{fitting the selected embedding model} 
\State $F_l = F_{l-1} - e_l.predict(t)$ \Comment{residual computation}
\If {$|F_l-F_{l-1}|<\epsilon:$} \Comment{Check termination condition}
    \State \bf{break}
\EndIf
\State $f^m.append(e_l)$
\EndFor
\State\Return $f^m$
\EndFunction
\end{algorithmic}
\caption{Boosted Embeddings Algorithm}
\label{algo:BEA}
\end{algorithm}

\noindent\paragraph{Drift Detection with Boosted Embeddings}
The proposed \texttt{UTDD} algorithm~\ref{algo:UTDD} is an unsupervised temporal data drift detection algorithm where we followed a sequence of steps to deseasonalize the time-series data. In line[3], we estimate the number of differences required to make the given time series stationary using Augmented Dickey–Fuller Functions. We observe from our experiments that the $max.diff=4$ is accurate for our domain. In line[4], we apply the estimated differencing.  

\begin{algorithm}
\begin{algorithmic}[1]
\small
\Function{UnsupervisedTemporalDriftDetector}{$\mathbf{X}$,$N$}
\State where $\mathbf{X}$ = $[X_1,X_2,..,X_{n=N}]$ such that $X_n \in \mathbb{R}$, $X_n$ is the datapoint at time $t$
\State $k = \func{ndiffs}(x, max.diff=4, test=``adf")$ \Comment{get min \# of differences estimates required to make a time series stationary}
\State $X$ = $X.diff(k)$
\State $P$ = $\func{BoostedEmbeddings}(\mathbf{X},N)$ 
\Comment{Fit BE to seasonality + additional categorical variables}
\State $Error = X-P$ \Comment{Compute residual}
\State $ZScoreCurr = \func{ComputeZScore}(Error)$
\State\Return $ZScoreCurr$
\EndFunction
\end{algorithmic}
\caption{Unsupervised Temporal Drift Detector}
\label{algo:UTDD}
\end{algorithm}
 
In line[5], we employ BE to fit to the seasonality along with any other additional categorical (that can be from domain expertise) in the resultant data. The BE model enables our approach to learn from the distinct time-series signature at once by encoding the categorical features (linear time, weekday or weekend effect, holiday effect, user-defined exogenous categorical variables etc) in a meaningful way, in a lower dimensional space to extract valuable information. We compute the residual, $Error$, by removing the seasonal component from the raw data $X$. Finally, we compute the z\_score of this residual, otherwise referred to as resultant stationary series, and evaluate if it has changed from the $z\_score$ data on which we had trained our model previously. Consequently, our approach is the first to do temporal drift detection in the context of unsupervised time series data, enabling automated cost-efficient retraining when the data has drifted.

\section{Experimental Evaluation}
\label{sec:exp}
In order to simulate time series data with multiple seasonal components with different periodicities, let's denote the time series by $x_t$, and the seasonal component $i$ by $\gamma_t^{(i)}$. For simplicity assume we just have two seasonal components:
\[x_t = \mu_t + \gamma_t^{(1)} + \gamma_t^{(2)} + \epsilon_t\]
where $\mu_t$ represents the trend or level, $\gamma_t^{(1)}$ represents a seasonal component with a relatively short period, and $\gamma_t^{(2)}$ represents another seasonal component of longer period and $\epsilon_t$ is white noise. Following equations (3.7) and (3.8) in Durbin and Koopman \cite{durbin2012time}, we can simulate the seasonality component $i$ with seasonal length of $s$ by
\begin{align}
\gamma^{(i)}_{t} &= \sum_{j=1}^p \gamma^{(i)}_{j, t} \\
\gamma^{(i)}_{j, t+1} &= \gamma^{(i)}_{j, t}\cos(\lambda_j) + \gamma^{*, (i)}_{j, t}\sin(\lambda_j) + \omega^{(i)}_{j,t},\\
\gamma^{*, (i)}_{j, t+1} &= -\gamma^{(i)}_{j, t}\sin(\lambda_j) + \gamma^{*, (i)}_{j, t}\cos(\lambda_j) + \omega^{*, (i)}_{j, t},
\end{align}
where $p=\lfloor s/2 \rfloor$ is the periodicity, $1\leq j \leq p$, $\lambda_j = \frac{2\pi j}{s}$, $\omega^{(i)}_{j,t}\sim N(0,\sigma^2_\omega)$ and $\omega^{*,(i)}_{j,t}\sim N(0,\sigma^2_{\omega^*})$.

\begin{figure*}[ht!]
	\begin{minipage}{0.23\linewidth}
			\includegraphics [width = \textwidth]{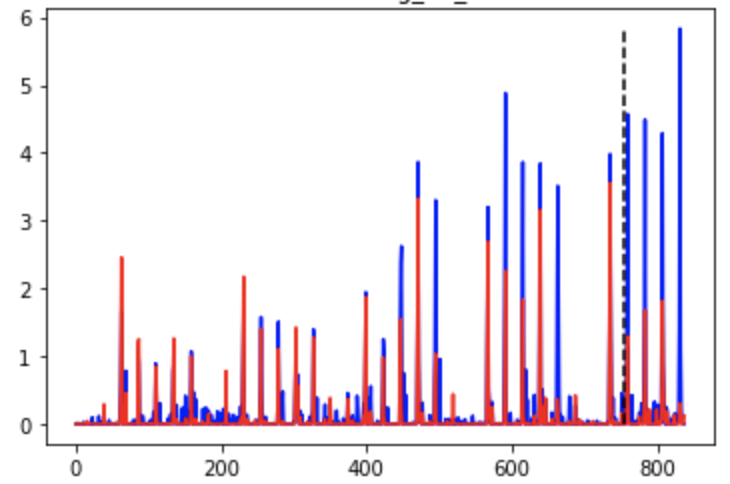}
			\caption{\small \centering Fitted BE to the seasonality to the data between Aug to Sep. Seasonality in Red. Residual in Blue.}
			\label{aug2Sep}
	\end{minipage}
	\begin{minipage}{0.23\linewidth}
		\centering
		\includegraphics [width = \textwidth]{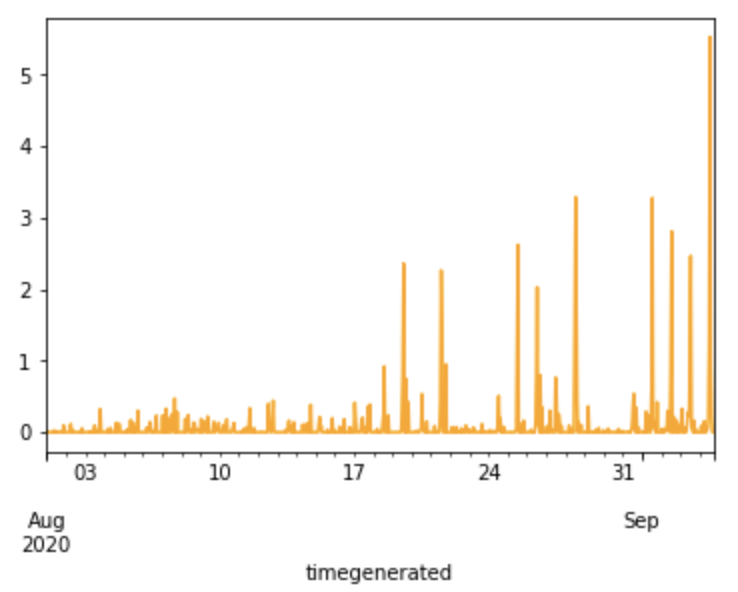}
		\caption{\small \centering Residual in Yellow for Aug to Sep. Residual Series will be used to compute z score.}
		\label{aug2Sep_residual}
	\end{minipage}
	\begin{minipage}{0.23\linewidth}
			\includegraphics [width = \textwidth]{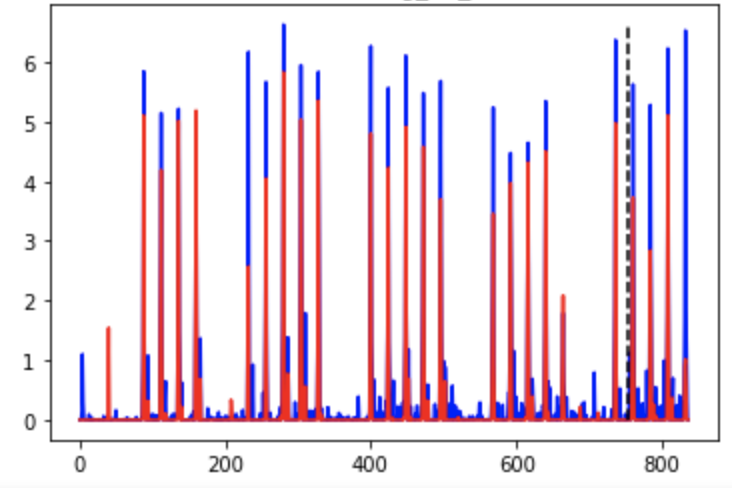}
			\caption{\small \centering Fitted BE to the seasonality to the data between Sep to Oct. Seasonality in Red. Residual in Blue.}
			\label{sep2oct}
	\end{minipage}
	\begin{minipage}{0.23\linewidth}
		\centering
		\includegraphics [width = \textwidth]{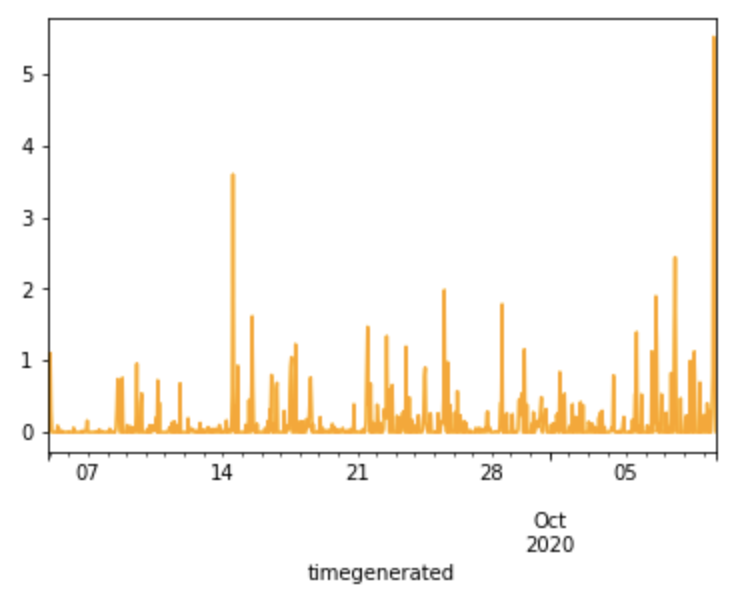}
		\caption{\small \centering Residual in Yellow for Sep to Oct. Residual Series will be used to compute z score.}
		\label{sep2oct_residual}
	\end{minipage}
	
\end{figure*}

We generated artificial data with seasonal patterns that differ in
their peaks and shapes during weekdays and weekends as well as holidays as shown in Fig~\ref{fig:train} from August through October. We estimate the drift between the data in August to September, and September to October respectively. To this effect, Fig~\ref{aug2Sep} is the result of fitting BE to the seasonality in data between Aug to Sep. Fig~\ref{aug2Sep_residual} is the residual as a result of applying line[6] of algorithm~\ref{algo:UTDD}. We repeat the same with the data in the range between Sep to Oct, such that, Fig~\ref{sep2oct} after fitting BE to the seasonality and Fig~\ref{sep2oct_residual} is the residual. Such that the resultant $z\_score$ of $0.53$ and $0.65$ for Fig~\ref{aug2Sep_residual} \& Fig~\ref{sep2oct_residual} respectively show that the data has drifted, as evident from the plots.

\section{Conclusion}
\label{sec:con}
The literature for concept drift detection mostly relies on  discovering significant drop in model generalization. In this paper, we follow the hypothesis that it is too unrealistic to assume that the labels are readily available as with the case with many real time problems. Therefore, we tackle the concept drift detection problem in an unsupervised manner. Unsupervised temporal drift detection methods are widely used in a variety of research areas as well as practical application domains. In this paper, we propose a novel label-free drift detection algorithm, \texttt{UTDD}, for time series data. In our knowledge, this would be the first work to do temporal drift detection in the context of unsupervised time series data by subtracting the effect of seasonality and modeling for the residual. 
Future work will focus on extending this preliminary study to evaluate and compare highlighted methods in related section by carrying out extensive experiments. 

\bibliographystyle{abbrv}
\bibliography{references}

\end{document}